%% file: main.tex
\documentclass[10pt,twocolumn,letterpaper]{article}

\usepackage[pagenumbers]{cvpr} 
\usepackage{url}            
\usepackage{booktabs}       
\usepackage{amsfonts}       
\usepackage{nicefrac}       
\usepackage{microtype}      
\usepackage{xcolor}         

\usepackage{adjustbox}
\usepackage{diagbox}
\usepackage{multirow}
\usepackage{makecell}

\usepackage{times}
\usepackage{epsfig}
\usepackage{graphicx}
\usepackage{amsmath}
\usepackage{amssymb}
\usepackage{colortbl}
\usepackage{xspace}

\usepackage{ragged2e}
\usepackage{float}
\usepackage{bbding}
\usepackage{amssymb}  
\usepackage[accsupp]{axessibility}
\usepackage{bbding}

\definecolor{cvprblue}{rgb}{0.21,0.49,0.74}
\usepackage[pagebackref,breaklinks,colorlinks,allcolors=cvprblue]{hyperref}

\def\paperID{***} 
\def\confName{CVPR}
\def\confYear{2025}

\title{Lift3D Foundation Policy: Lifting 2D Large-Scale Pretrained Models for Robust 3D Robotic Manipulation}

\author{
Yueru Jia\textsuperscript{\rm 1,2$^{*}$}, Jiaming Liu\textsuperscript{\rm 1$^{*,\dagger}$}, Sixiang Chen\textsuperscript{\rm 1$^{*}$}, Chenyang Gu\textsuperscript{\rm 1}, Zhilue Wang\textsuperscript{\rm 1}, Longzan Luo\textsuperscript{\rm 1}, 
\\Lily Lee\textsuperscript{\rm 1}, Pengwei Wang\textsuperscript{\rm 2}, Zhongyuan Wang\textsuperscript{\rm 2}, Renrui Zhang\textsuperscript{\rm 3$^{\dagger}$}, Shanghang Zhang\textsuperscript{\rm 1,2}~\textsuperscript{\Envelope}
\vspace{0.2cm}\\
\textsuperscript{\rm 1}State Key Laboratory of Multimedia Information Processing, School of Computer Science,\\Peking University; 
\textsuperscript{\rm 2}Beijing Academy of Artificial Intelligence (BAAI); \textsuperscript{\rm 3}CUHK\\
$^{*}$ Equal contribution, $^{\dagger}$ Project lead, \Envelope 
 Corresponding author\\
\textbf{Project web page:} \href{https://lift3d-web.github.io/}{lift3d-web.github.io}
}

\begin{document}
\twocolumn[
{%
\renewcommand\twocolumn[1][]{#1}
\maketitle
\begin{center}
\centering
\begin{minipage}[t]{\linewidth}
\includegraphics[width=\textwidth]{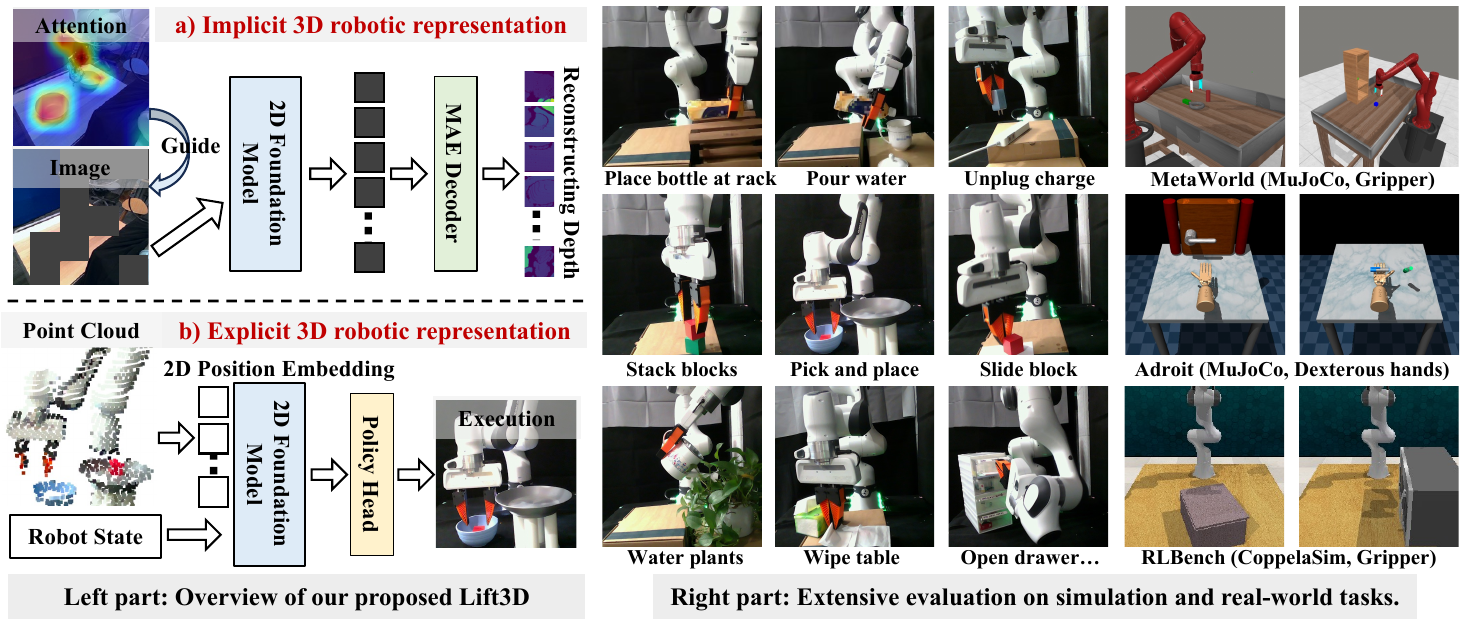}
\vspace{-0.6cm}
{\captionsetup{hypcap=false}  
\captionof{figure}{\footnotesize{\textbf{
Lift3D empowers 2D foundation models with 3D manipulation capabilities by refining implicit 3D robotic representations through task-related affordance masking and depth reconstruction, while enhancing explicit 3D robotic representations by leveraging the pretrained 2D positional embeddings to encode point cloud. Lift3D achieves robustness and surprising effectiveness in diverse simulation and real-world tasks.
}}}
\label{fig:teaser}}
\end{minipage}
\end{center}
}]

\input{sec/0_abstract}    
\input{sec/1_intro}

\input{sec/2_formatting}
\input{sec/3_experiment}

{
    \small
    \bibliographystyle{ieeenat_fullname}
    \bibliography{main}
}

\clearpage

\input{sec/Appendix}

\end{document}

%% file: sec/0_abstract.tex
\begin{abstract}

3D geometric information is essential for manipulation tasks, as robots need to perceive the 3D environment, reason about spatial relationships, and interact with intricate spatial configurations. Recent research has increasingly focused on the explicit extraction of 3D features, while still facing challenges such as the lack of large-scale robotic 3D data and the potential loss of spatial geometry. To address these limitations, we propose the Lift3D framework, which progressively enhances 2D foundation models with implicit and explicit 3D robotic representations to construct a robust 3D manipulation policy. Specifically, we first design a task-aware masked autoencoder that masks task-relevant affordance patches and reconstructs depth information, enhancing the 2D foundation model's implicit 3D robotic representation. After self-supervised fine-tuning, we introduce a 2D model-lifting strategy that establishes a positional mapping between the input 3D points and the positional embeddings of the 2D model. Based on the mapping, Lift3D utilizes the 2D foundation model to directly encode point cloud data, leveraging large-scale pretrained knowledge to construct explicit 3D robotic representations while minimizing spatial information loss. In experiments, Lift3D consistently outperforms previous state-of-the-art methods across several simulation benchmarks and real-world scenarios.

\end{abstract}

%% file: sec/1_intro.tex
\section{Introduction}
\label{sec:intro}

A fundamental goal of vision-based manipulation policies is to understand the scene and predict corresponding 3D poses. Some existing methods utilize 2D images as input to directly predict 3D end-effector poses through reinforcement learning~\cite{yarats2021mastering, lobbezoo2021reinforcement, chen2022towards, geng2023rlafford, an2024rgbmanip, hu2023imitation} or imitation learning~\cite{kim2024openvla, liu2024robomamba, li2023manipllm, chi2023diffusion, zitkovich2023rt, brohan2022rt, zhao2023learning, fang2019survey}. 
While these approaches can effectively handle a range of manipulation tasks, they fall short of fully understanding the spatial relationships and 3D structures in the physical world~\cite{shridhar2022cliport, zhu2024point, eisner2022flowbot3d, fang2023anygrasp, shridhar2023perceiver}.
In robotic manipulation, 3D geometric information is crucial for tackling complex tasks, as robots must perceive the 3D environment, reason about geometric relationships, and interact with intricate spatial configurations.

Recent research has increasingly focused on the explicit extraction of 3D feature representations in robotic manipulation tasks, which can be categorized into two groups.
\textbf{On one hand, some methods directly encode point cloud data}~\cite{shridhar2023perceiver, chen2023polarnet, liu2022frame, zhang2024leveraging, wang2024rise, ze20243d, james2022coarse}, either training 3D policy models from scratch or fine-tuning pretrained point cloud encoders(i.e., PointNet++~\cite{qi2017pointnet++} and PointNext~\cite{qian2022pointnext}). However, the limited availability of large-scale robotic 3D data and foundational models constrains their generalization capabilities. Additionally, processing 3D or voxel features incurs significant computational costs, hindering scalability and practicality in real-world applications.
\textbf{On the other hand, some methods involve transforming modalities}, such as lifting pretrained 2D features into 3D space~\cite{gervet2023act3d, ke20243d, xian2023chaineddiffuser, shridhar2022cliport} or projecting 3D point clouds into multi-view images for input into 2D pretrained models~\cite{goyal2023rvt, goyal2024rvt, wang2024vihe, zhang2024sam}.
Despite showing promising performance on several downstream manipulation tasks, these modality transformations inevitably lead to a loss of spatial information, hindering robots' ability to understand 3D spatial relationships.
Building on the challenges in the aforementioned 3D policies, we raise a question: ``\textit{Can we develop a 3D policy model that integrates large-scale pretrained knowledge while incorporating complete 3D spatial data input?}"

To address this question, we propose the Lift3D framework, which elevates transformer-based 2D foundation model (e.g., DINOV2~\cite{oquab2023dinov2} or CLIP~\cite{radford2021learning}) to construct robust 3D manipulation policy step by step. The key insight of Lift3D is first to enhance the implicit 3D robotic representation, followed by the explicit encoding of point cloud data for policy imitation learning.
For implicit 3D robotic representation, we design a task-aware masked autoencoder (MAE) that processes 2D images and reconstructs 3D geometric information in a self-supervised manner, as shown in Figure~\ref{fig:teaser} \textcolor{red}{a)}. 
Specifically, we utilize large-scale unlabeled datasets from robotic manipulation~\cite{o2023open, grauman2022ego4d} and leverage a multimodal model (i.e., CLIP) to extract image attention maps based on task text descriptions. 
These attention maps are then back-projected onto the 2D input to guide the MAE masking strategy, focusing on task-related affordance regions.
Reconstructing the depth of masked tokens enhances the 3D spatial awareness of the 2D foundation model, facilitating subsequent point cloud imitation learning.

For explicit 3D robotic representation, we propose a 2D model-lifting strategy that directly leverages a 2D foundation model to encode 3D point cloud data, as shown in Figure~\ref{fig:teaser} \textcolor{red}{b)}. Specifically, inspired by the virtual camera setting in~\cite{goyal2023rvt, goyal2024rvt}, we first project the point cloud data onto multiple virtual planes. 
However, our projection process is not aimed at constructing the input for the policy model, but rather at establishing a positional correspondence between the input 3D points and the pretrained 2D positional embeddings (PEs) of each virtual plane.
Guided by this positional mapping, the 2D foundation model can use its original PEs to encode the point cloud data, enabling the model to extract 3D features based on its large-scale pretrained knowledge. Unlike previous methods, Lift3D eliminates modality transformation during imitation learning, minimizing the loss of robotic spatial information, while reducing computational cost by directly leveraging the 2D foundation model for forward propagation. Through a two-stage training process, Lift3D enhances the 2D foundation model with robust 3D robotic manipulation capabilities by systematically improving both implicit and explicit 3D robotic representations.

To comprehensively evaluate our proposed Lift3D, we conduct extensive experiments across three simulation benchmarks~\cite{yu2020meta, rajeswaran2017learning, james2020rlbench} and several real-world scenarios, including over 30 different gripper and dexterous hand manipulation tasks, as shown in Figure~\ref{fig:teaser}.
We compare various baseline methods, such as robotic 2D representation methods~\cite{majumdar2023we, nair2022r3m, radford2021learning}, 3D representation methods~\cite{zhu2024spa, qi2017pointnet++, qian2022pointnext}, and 3D imitation learning policies~\cite{goyal2024rvt, ke20243d, goyal2023rvt}. 
Lift3D consistently outperforms other methods, even when utilizing the simplest MLP policy head and single-view point cloud, demonstrating the robustness of our model’s manipulation capabilities and its understanding of robotic 3D spatial awareness. 
For instance, Lift3D achieves improvements in the average success rates of 18.2\% and 21.3\% over previous state-of-the-art 3D policy methods~\cite{ze20243d} on the Meta-World and Adroit benchmarks, respectively. 
We also explore the model scalability across several complex tasks, gradually increasing the parameters of the 2D foundation model.
In real-world experiments, Lift3D can learn novel manipulation skills with just 30 episodes per task.
To evaluate the generalization capabilities of Lift3D, we incorporate different manipulated instances, background scenes, and lighting conditions from the training set into the real-world testing process.
Lift3D shows strong generalization abilities, effectively leveraging the large-scale pretrained knowledge of the 2D foundation model and comprehensive 3D robotic representations.
In summary, our contributions are as follows:

\begin{itemize}
\item 
We propose Lift3D, which elevates 2D foundation models to construct a 3D manipulation policy by systematically improving implicit and explicit 3D robotic representations.

\item 
For implicit 3D robotic representation, we design a task-aware MAE that masks task-related affordance regions and reconstructs the depth geometric information, enhancing the 3D spatial awareness of the 2D foundation model.

\item 
For explicit 3D robotic representation, we propose a 2D model-lifting strategy that leverages the pretrained PEs of a 2D foundation model to encode 3D point cloud data for manipulation imitation learning.

\end{itemize}

%% file: sec/2_formatting.tex
\section{Related work}

\textbf{Robotic Representation Learning}
In recent years, substantial progress has been made in the field of pretrained visual representations, with a primary focus on self-supervised learning paradigms such as contrastive learning~\cite{he2020momentum, chen2021empirical, chen2020simple, oquab2023dinov2, liu2024unsupervised}, self-distillation~\cite{baevski2022data2vec, caron2020unsupervised}, and masked autoencoders (MAE)~\cite{he2022masked, bachmann2022multimae, xie2022simmim, fan2024text, wang2023videomae, zhang2023learning, liu2024continual}. Additionally, multi-modal alignment approaches~\cite{radford2021learning, zhai2023sigmoid} leverage large-scale paired image-language data to learn more semantically-aware representations.
Building on the above research, more studies are dedicated to enhancing 2D representations in the field of robotics.
Specifically, R3M~\cite{nair2022r3m} leverages contrastive learning to learn a universal embodied representation from diverse human video data.
Vip~\cite{ma2022vip} generates dense, smooth reward functions for unseen robotic tasks.
MVP~\cite{radosavovic2023real}, VC-1~\cite{majumdar2023we}, and Voltron~\cite{karamcheti2023language} delve into the effectiveness of the MAE strategy in robotic pretraining.
~\cite{zeng2024learning,yang2024spatiotemporal} consider the impact of multiple frames on the current state. 
However, 2D representations often struggle to capture the spatial context needed for complex robotic tasks, prompting other pretraining work to explore 3D robotic representations. 
MV-MWM ~\cite{seo2023multi}, 3D-MVP~\cite{qian20243d}, and SPA~\cite{zhu2024spa} leverage multi-view MAE to learn 3D visual representation. SUGAR~\cite{chen2024sugar} and Point Cloud Matters~\cite{zhu2024point} introduce point cloud-based 3D representations, showing that these observations often improve policy performance and generalization capabilities. DPR~\cite{wang2024visual} leverages depth information as auxiliary knowledge for pretraining.
Unlike previous robotic pretraining methods, Lift3D fully leverages existing 2D foundation models~\cite{oquab2023dinov2, radford2021learning} by enhancing them with implicit 3D robotic representations. This approach makes the first attempt to mask task-related affordance regions and reconstruct depth information, not only improving 3D spatial awareness but also facilitating subsequent point cloud imitation learning.

\textbf{Robotic Manipulation.}
Traditional robotic manipulation typically relies on state-based reinforcement learning~\cite{andrychowicz2020learning, geng2022end, joshi2020robotic, yarats2021mastering}. In contrast, recent approaches~\cite{eisner2022flowbot3d, wang2023sparsedff, mo2021where2act, chi2023diffusion, zhao2023learning, li2023imagemanip} utilize visual observations as input to inform predictions.
Inspired by the success of Multimodal Large Language Models~\cite{li2023blip2, awadalla2023openflamingo, guo2023point, zhang2024mavis, yang2023lidar}, several Vision-Language-Action models~\cite{driess2023palm, zitkovich2023rt, li2023vision, huang2023voxposer, li2023manipllm, liu2024robomamba, liu2024self} are designed for task planning and manipulation. 
However, ~\cite{zhu2024point} demonstrates that explicit 3D representation is essential for complex robotic tasks, offering strong robustness to variations.
Another category of works focuses on directly encoding 3D information to predict end-effector poses~\cite{liu2022frame, zhang2024leveraging, wang2024rise, ze20243d, james2022coarse}.
Anygrasp~\cite{fang2023anygrasp} employs point cloud data to learn grasp poses on a large-scale dataset.
~\cite{goyal2023rvt, goyal2024rvt, wang2024vihe} project 3D point clouds into multi-view images for input into large-scale 2D pretrained models, while ~\cite{gervet2023act3d, ke20243d, xian2023chaineddiffuser, shridhar2022cliport} lift pretrained 2D features into 3D space.
PolarNet~\cite{chen2023polarnet} and M2T2~\cite{yuan2023m2t2} directly utilize point clouds reconstructed from RGB-D data, processing them through an encoder-transformer combination for action prediction, while C2F-ARM~\cite{james2022coarse} and PerAct~\cite{shridhar2023perceiver} employ voxelized point clouds with a 3D backbone for action inference. Act3D~\cite{gervet2023act3d} and ChainedDiffuser~\cite{xian2023chaineddiffuser} take a different approach by representing scenes as multi-scale 3D features.
3D Diffusion Actor~\cite{ke20243ddiffuseractorpolicy} and 3D Diffusion Policy~\cite{chi2023diffusion} harness diffusion models to generate precise actions in 3D space.
Unlike previous 3D policies, Lift3D directly encodes point cloud data using a pretrained 2D foundation model and its positional embeddings, minimizing spatial information loss and enhancing the generalization capability.

\begin{figure*}[t]
\centering
\includegraphics[width=\linewidth]{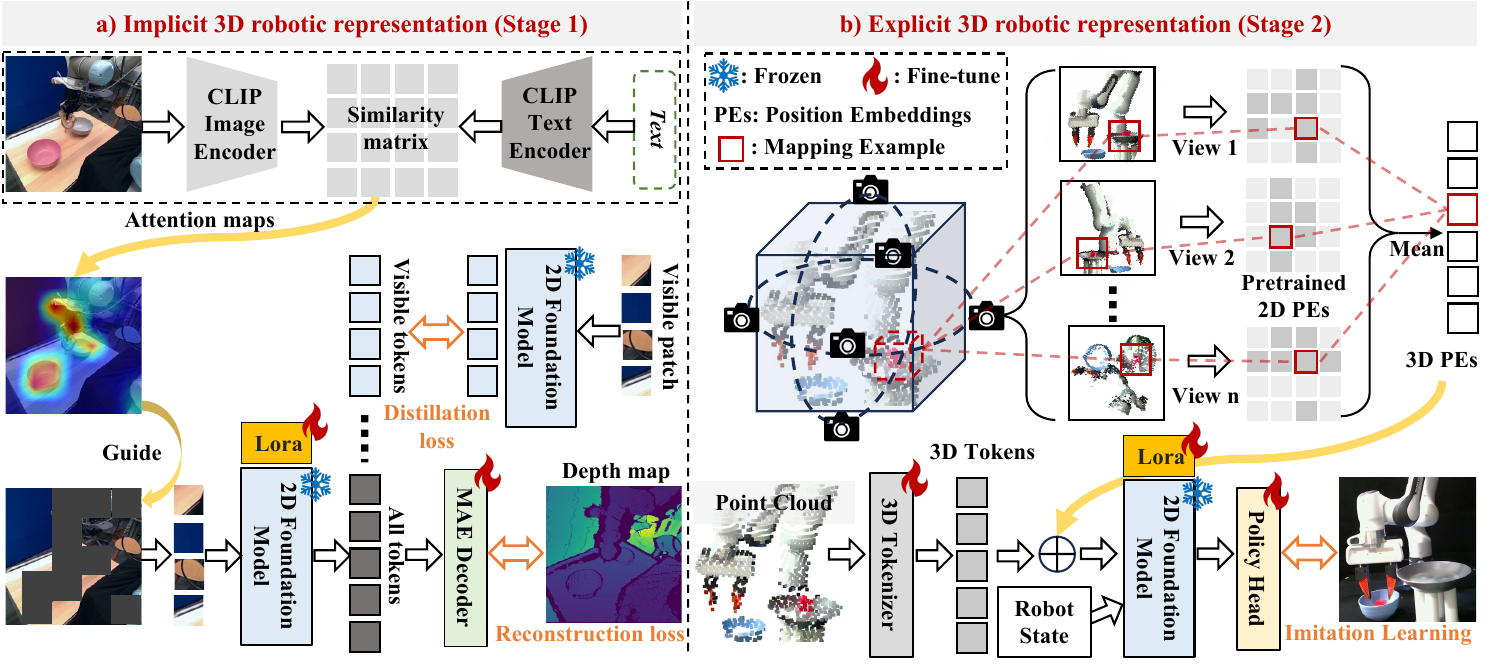}
\vspace{-0.7cm}
\caption{\textbf{Overall pipeline of Lift3D.} 
\textcolor{red}{a) For implicit 3D robotic representation}, we leverage CLIP~\cite{radford2021learning} to offline extract image attention maps based on task descriptions, which are back-projected onto the 2D input to guide the MAE masking. 
We then input the visible tokens into the 2D foundation model to extract features. The masked tokens and encoded visible tokens are processed by the MAE decoder for depth reconstruction, enhancing 3D spatial awareness. Meanwhile, the encoded visible tokens are also distilled using corresponding features from the off-the-shelf pretrained model to mitigate catastrophic forgetting.
\textcolor{red}{b) For explicit 3D robotic representation}, we first project the point cloud data onto multiple virtual planes, establishing a positional mapping between the 3D input points and the 2D positional embeddings (PEs) on each virtual plane. 
After mapping, we average the 2D PEs corresponding to each 3D patch to form a unified positional indicator(3D PEs), which is then integrated with the 3D tokens.
These 3D tokens are generated by feeding the point cloud into a lightweight 3D tokenizer. Finally, the output features from the 2D foundation model are processed through a policy head to predict the pose for imitation learning.
}
\label{fig:framework} 
\vspace{-0.3cm}
\end{figure*}

\section{Lift3D Method}
In Section \ref{sec:PS}, we introduce the problem statement of our proposed Lift3D framework. In Sections \ref{sec:TMA} and \ref{sec:2MS}, we detail the technical aspects of our Task-aware MAE and 2D Model-lifting Strategy, which enhance implicit and explicit 3D robotic representations, respectively

\subsection{Problem Statement}
\label{sec:PS}
\textbf{For implicit 3D robotic representation}, following previous MAE methods~\cite{he2022masked, radosavovic2023real}, we first input the masked image $I \in \mathbb{R}^{W \times H \times 3}$ into the 2D foundation model $2D_e$. The output features are then fed into a decoder $2D_d$ for depth reconstruction, $D = 2D_d(2D_e(I))$, where $D \in \mathbb{R}^{W \times H \times 1}$. This process enhances the 3D spatial awareness of the $2D_e$ model and facilitates subsequent 3D imitation learning.

\noindent \textbf{For explicit 3D robotic representation}, we directly utilize the 2D foundation model $2D_e$ to encode 3D point cloud data $P \in \mathbb{R}^{N \times 6}$ and the robot state $R_S$. 
We then use a simple policy head $\pi$ to predict the action $a = \pi(2D_e(P, R_S))$. Following previous manipulation works~\cite{goyal2023rvt, gervet2023act3d}, we adopt 7-DoF action to express the end-effector pose of the robot arm, which includes 3-DoF for translation, 3-DoF for rotation, and 1-DoF for the gripper state (open or closed).

\subsection{Task-aware Masked Autoencoder}
\label{sec:TMA}
Several studies~\cite{khandelwal2022simple, majumdar2023we, shang2024theia} have shown that 2D foundation models exhibit strong representation and generalization capabilities across various downstream robotic tasks. Building on this, Lift3D first enhances the implicit 3D robotic representations within 2D foundation models.
Existing robotic MAE reconstruction methods~\cite{radosavovic2023real, majumdar2023we, qian20243d} employ an aggressive masking strategy, where a large portion of the input image patches is randomly masked. However, the masked patches may mostly contain irrelevant background information, hindering the effective learning of foreground object representations. 
Unlike previous methods, Lift3D aims to mask task-related affordance patches and reconstruct the depth geometric information, enhancing the 3D spatial awareness of the 2D foundation model.
Specifically, we leverage large-scale datasets from robotic manipulation~\cite{o2023open} to construct our MAE training dataset, which includes 1 million training samples randomly sampled from videos, consisting of paired image and depth data.
The additional details of the reconstruction dataset are shown in Appendix~\ref{apsec: AD}.
As shown in Figure~\ref{fig:framework} a), once the data is obtained, we use a multimodal model (i.e., CLIP) to generate image attention maps based on task-specific text descriptions.
For instance, the text prompt used to extract the attention map in Figure~\ref{fig:framework} is "Robot arm take the red bowl and put it in the grey bowl." These attention maps are then bilinearly resized and back-projected onto the input image to guide the MAE masking strategy. To distinguish task-related affordance tokens from background tokens, we apply a threshold (i.e., $\theta = 0.5$) to filter the attention values of all tokens. Note that the attention value for each token is computed by averaging its pixel-level values.
Consistent with the masking ratio used in previous methods~\cite{he2022masked}, we also randomly mask background tokens to achieve the desired ratio (i.e., $r = 75.0\%$).

The visible ($x_{\text{vis}}$) tokens are fed into the 2D foundation model, then concatenated with the masked tokens ($x_{\text{mask}}$), which are directed to a decoder for reconstruction.
The reconstruction target plays a crucial role in masked image modeling, directly influencing the learning of feature representations. Previous robotic MAE methods typically use low-level RGB information~\cite{majumdar2023we, qian20243d, radosavovic2023real} as the reconstruction target.
To enhance the 3D spatial awareness of the 2D foundation model, we reconstruct depth information ($D_{\text{target}}$) for task-related affordance patches and randomly selected background patches.
Finally, to preserve the inherent capabilities of the foundation model, we introduce a distillation loss that constrains the distance between our model's visible token outputs and the corresponding features from the off-the-shelf pretrained model ($2D_{e}^{\text{pre}}$).
As shown in Figure~\ref{fig:framework} a), during the stage 1 training process, we fine-tune the injected adapter~\cite{hu2021lora} and decoder using reconstruction and distillation losses, which are formulated as:
\begin{align}
\mathcal{L}_{\text{implicit}} = & \left\| 2D_{e}(x_{\text{vis}}) - 2D_{e}^{\text{pre}}(x_{\text{vis}}) \right\|_1 
 + \\ &\left\| 2D_{d}\left(2D_{e}(x_{\text{vis}}) \parallel x_{\text{mask}} \right) - D_{\text{target}} \right\|_1
\end{align}

\subsection{2D Model-lifting Strategy}
\label{sec:2MS}
After endowing the 2D foundation model with implicit 3D robotic awareness, we introduce a lifting strategy that enables the 2D model to explicitly comprehend point cloud data.
Recent works, whether projecting 3D point clouds into multi-view images~\cite{goyal2023rvt, wang2024vihe} or lifting 2D features into 3D space~\cite{gervet2023act3d, shridhar2022cliport}, have faced the challenge of losing spatial information due to modality transformation. Therefore, efficiently encoding 3D data has been a key research focus in the 3D robotics field.
For transformer-based 2D models, positional embeddings (PEs) play an important role, as they provide positional indicators for input tokens within the attention mechanism.
However, directly creating new 3D PEs to encode 3D tokens could introduce semantic discrepancies between the pretrained 2D foundation model and the newly added 3D PEs, potentially causing a loss of large-scale pretrained knowledge.

Therefore, inspired by ~\cite{tang2025any2point, goyal2024rvt, goyal2023rvt}, we project the 3D tokens onto multiple virtual planes. Unlike previous works, our projection process is not intended to construct the model's input. Instead, it establishes a positional correspondence between the input 3D points and the pretrained 2D PEs of each virtual plane. These 2D PEs are then used to directly encode the 3D tokens.
Specifically, as shown in Figure~\ref{fig:framework} b), we transform the raw point clouds into high-dimensional space (i.e., $B \times 128 \times 768$), obtaining $k$ (128) 3D tokens through a lightweight 3D tokenizer, following previous methods~\cite{zhang2023starting, zhu2024no}.
The 3D tokenizer consists of farthest point sampling~\cite{qi2017pointnet} for downsampling the number of points, the k-Nearest Neighbor algorithm for local aggregation, and learnable linear layers for feature encoding. Additionally, we use $\{C_{3D}^{i} \}_{i=1}^{k}$ to represent the 3D coordinates of each 3D token. 
Subsequently, each 3D coordinate are projected onto $n$ virtual planes, obtaining corresponding 3D-to-2D coordinates $\{C_{2D}^{ij}\}_{j=1}^n$. The projection mechanism is parameter-free and efficient. We adopt a cube projection method with 6 faces, effectively capturing spatial information.
The $n$ virtual planes correspond to $n$ original 2D PEs.
Using the 3D-to-2D coordinates, we assign each 3D token to $n$ original 2D PEs, denoted as $\{{PE}_{2D}(C_{2D}^{ij})\}_{j=1}^n$.

After aligning each 3D token with $n$ 2D PEs, we simply average them to create a unified positional indicator, denoted as $PE_{3D}$, which is formulated as:
\begin{align}
\vspace{-0.3cm}
PE_{3D} = \frac{1}{n}\sum_{j=1}^n (PE_{2D}(C_{2D}^{ij}))
\vspace{-0.3cm}
\end{align}
We incorporate $PE_{3D}$ with the 3D tokens and input them into the 2D foundation model. In this way, we utilize the $n$ combined original 2D PEs to encode the 3D tokens, which effectively provides diverse positional relations within 2D space and mitigates spatial information loss.
The output features (i.e., $B \times 128 \times 768$) from the 2D foundation model are processed through a simple policy head to predict the pose for imitation learning.
We employ a three-layer multilayer perceptron (MLP) to construct the policy head. It is important to note that our Lift3D encoder can be easily adapted to different decoders or policy heads; the MLP head is used here for simple validation.
Finally, the supervision loss is formulated as:
\begin{align}
\mathcal{L}_{\text{explicit}} = & \ \text{MSE}(T_{\text{pred}}, T_{\text{gt}}) + \left(1 - \frac{R_{\text{pred}} \cdot R_{\text{gt}}}{\|R_{\text{pred}}\| \|R_{\text{gt}}\|}\right)  \\
& + \text{BCE}(G_{\text{pred}}, G_{\text{gt}})
\end{align}

Where T, R, and G represent translation, rotation, and gripper state in the 7-DoF end-effector pose, respectively. As shown in Figure~\ref{fig:framework} b), during stage 2 imitation learning, we freeze the parameters of the 2D foundation model and update only the 3D tokenizer, injected adapter, and policy head. Lift3D can also operate without injecting the adapter, resulting in a minor reduction in manipulation performance. The additional validations are shown in Appendix~\ref{apsec: AAS}.

%% file: sec/3_experiment.tex
\section{Experiments}

In Sections~\ref{sec: SE} and~\ref{sec: RE}, we evaluate the manipulation capability of our proposed Lift3D by presenting the experimental settings and results from both simulation and real-world tasks, respectively. The effectiveness of each component is validated through an ablation study in Section~\ref{sec: AS}. 
The generalization capabilities of Lift3D are examined in Section~\ref{sec: Ge}, where we test the model on different manipulated instances, background scenes, and lighting conditions.
In Section~\ref{sec: MS}, we explore the model scalability by gradually increasing the parameters of the 2D foundation model.

\input{tab/table_04_overall}

\subsection{Simulation Experiment}
\label{sec: SE}

\textbf{Benchmarks.}
We select over 30 tasks from three widely-used manipulation simulation benchmarks: MetaWorld~\cite{yu2020meta} and Adroit~\cite{rajeswaran2017learning} in the MuJoCo simulator, and RLBench~\cite{james2020rlbench} in the CoppeliaSim simulator. The point cloud data is derived from single-view RGBD data using the camera extrinsics and intrinsics.
For \textbf{MetaWorld}, a tabletop environment with a Sawyer arm and two-finger gripper, we select 15 tasks of varying difficulty levels~\cite{seo2023masked}. 
These tasks, captured from two corner camera perspectives, are categorized as follows: easy tasks include \textit{button-press}, \textit{drawer-open}, \textit{reach}, \textit{handle-pull}, \textit{peg-unplug-side}, \textit{lever-pull}, and \textit{dial-turn}; medium tasks include \textit{hammer}, \textit{sweep-into}, \textit{bin-picking}, \textit{push-wall}, and \textit{box-close}; hard and very hard tasks include \textit{assembly}, \textit{hand-insert}, and \textit{shelf-place}.
For \textbf{Adroit}, which focuses on dexterous hand manipulation with the same camera view as in~\cite{majumdar2023we}, we include three tasks: \textit{hammer}, \textit{door}, and \textit{pen}.
For \textbf{RLBench}, which uses a Franka Panda robot and a front-view camera. Due to space constraints, the results and details of RLBench are provided in Appendix~\ref{apsec: RE}.

\noindent \textbf{Data collection.}
Scripted policies are used in MetaWorld~\cite{yu2020meta}, where 25 demonstrations are collected, each consisting of 200 steps. For the Adroit tasks, trajectories are obtained from agents trained with reinforcement learning algorithms. Specifically, DAPG~\cite{rajeswaran2017learning} is applied to the \textit{door} and \textit{hammer} tasks, while VRL3~\cite{wang2022vrl3} is used for the \textit{pen} task. We collect 100 demonstrations, each consisting of 100 steps.
Demonstrations in RLBench are collected through pre-defined waypoints and the Open Motion Planning Library~\cite{sucan2012open}, with 100 episodes gathered, each containing several key frames.

\noindent \textbf{Baselines.} 
The innovation of Lift3D lies in systematically enhancing both implicit and explicit 3D robotic representations. To evaluate its effectiveness, we compare Lift3D with 9 methods from three categories:
\textbf{1) 2D Robotic Representation Methods.} We select CLIP (ViT-base)~\cite{radford2021learning}, which is a 2D foundation model. Additionally, R3M~\cite{nair2022r3m} and VC-1~\cite{majumdar2023we} are included, both of 2D robotic pretraining methods.
\textbf{2) 3D Representation Methods.} Following~\cite{zhu2024point}, we incorporate foundational 3D models, including PointNet~\cite{qi2017pointnet}, PointNet++~\cite{qi2017pointnet++}, and PointNext~\cite{qian2022pointnext}. Additionally, we examine SPA~\cite{zhu2024spa}, the previous SOTA 3D robotic pretraining method.
Following~\cite{majumdar2023we}, all robotic representation methods use the same three-layer policy head and training loss as Lift3D.
\textbf{3) 3D Policies.} Lift3D is compared with the previous SOTA 3D Diffusion Policy (DP3)~\cite{ze20243d} on MetaWorld and Adroit, and with RVT-2~\cite{goyal2024rvt} on RLBench.

\noindent \textbf{Training and Evaluation Details.} 
For fair comparisons, all baselines are trained and evaluated under the same configuration, while 3D policy methods follow their original settings. The 2D and 3D visual inputs consist of \(224 \times 224\) RGB images and single-view point clouds with 1,024 points, respectively. The robot state includes end-effector pose, joint positions, and velocities, which are concatenated with the visual features. We use CLIP~\cite{radford2021learning} and DINOV2~\cite{oquab2023dinov2} (ViT-base) as our 2D foundation models. Following previous work~\cite{majumdar2023we}, the Adam optimizer is employed with parameters $(\beta_1, \beta_2)=(0.9, 0.999)$ and a learning rate of 1e-3. A constant learning rate is used for MetaWorld, while a cosine annealing scheduler with a 0.1 warm-up factor is applied for Adroit. Each method is trained for 100 epochs, with 25 rollouts performed every 10 epochs. We report the average success rate of the best-performing policy models across training. For MetaWorld, we further average the scores across the two camera views.

\noindent \textbf{Quantitative Results.} 
In Table~\ref{tab:sim_exp}, Lift3D(CLIP) achieves an average success rate of 83.9 on the MetaWorld benchmark, with 78.8 accuracy on medium-level tasks and 82.0 accuracy on hard level tasks. Compared to other robotic representation methods, Lift3D improves the mean success rate by 8.8 over the top-performing 2D method and by 14.4 over the top-performing 3D method. 
In addition, compared to the previous SOTA 3D policy (DP3), Lift3D achieves an accuracy improvement of 18.6. These results demonstrate that Lift3D effectively enhances the 2D foundation model with robust manipulation capabilities, enabling a deeper understanding of robotic 3D scenes by leveraging large-scale pretrained knowledge.
Furthermore, Lift3D also achieves superior performance on dexterous hand tasks compared to previous robotic representation and policy methods. Note that the DoF of the dexterous hand varies across tasks, with the \textit{hammer}, \textit{door}, and \textit{pen} tasks having 26, 28, and 24 DoF, respectively. The results demonstrate that our method is also effective for more complex dexterous hand manipulation tasks, owing to the strong 3D robotic representation. 
Lift3D(DINOV2) also shows promising results, demonstrating the method's practicality for other 2D foundation models.
The detailed scores are shown in Appendix~\ref{apsec: DSM}.

\subsection{Real-World Experiment}
\label{sec: RE}

\textbf{Dataset collection.}
In our real-world setup, we conduct experiments using a Franka Research 3 arm, with a static front view captured by an Intel RealSense L515 RGBD camera. We perform ten tasks: 1) \textit{place bottle at rack}, 2) \textit{pour water}, 3) \textit{unplug charger}, 4) \textit{stack blocks}, 5) \textit{pick and place}, 6) \textit{slide block}, 7) \textit{water plants}, 8) \textit{wipe table}, 9) \textit{open drawer}, and 10) \textit{close drawer}. These tasks involve various types of interacted objects and manipulation actions.
For each task, 40 demonstrations are collected in diverse spatial positions, with trajectories recorded at 30 fps. We select 30 episodes and extract key frames to construct the training set for each task. Examples of input point cloud data and images are displayed in Figure~\ref{fig:realvis}.
Additional details of the real-world dataset and assets are provided in Appendix~\ref{apsec: AD}.

\noindent \textbf{Training and Evaluation Details.}
The implementation details are the same as in our simulation experiments. We train each method from scratch for each task. During the training process, point clouds and action poses in the world coordinate system are used as inputs and supervision, respectively. For evaluation, we use the model from the final epoch and evaluate it 20 times in diverse spatial positions.

\begin{figure}[t]
\includegraphics[width=0.475\textwidth]{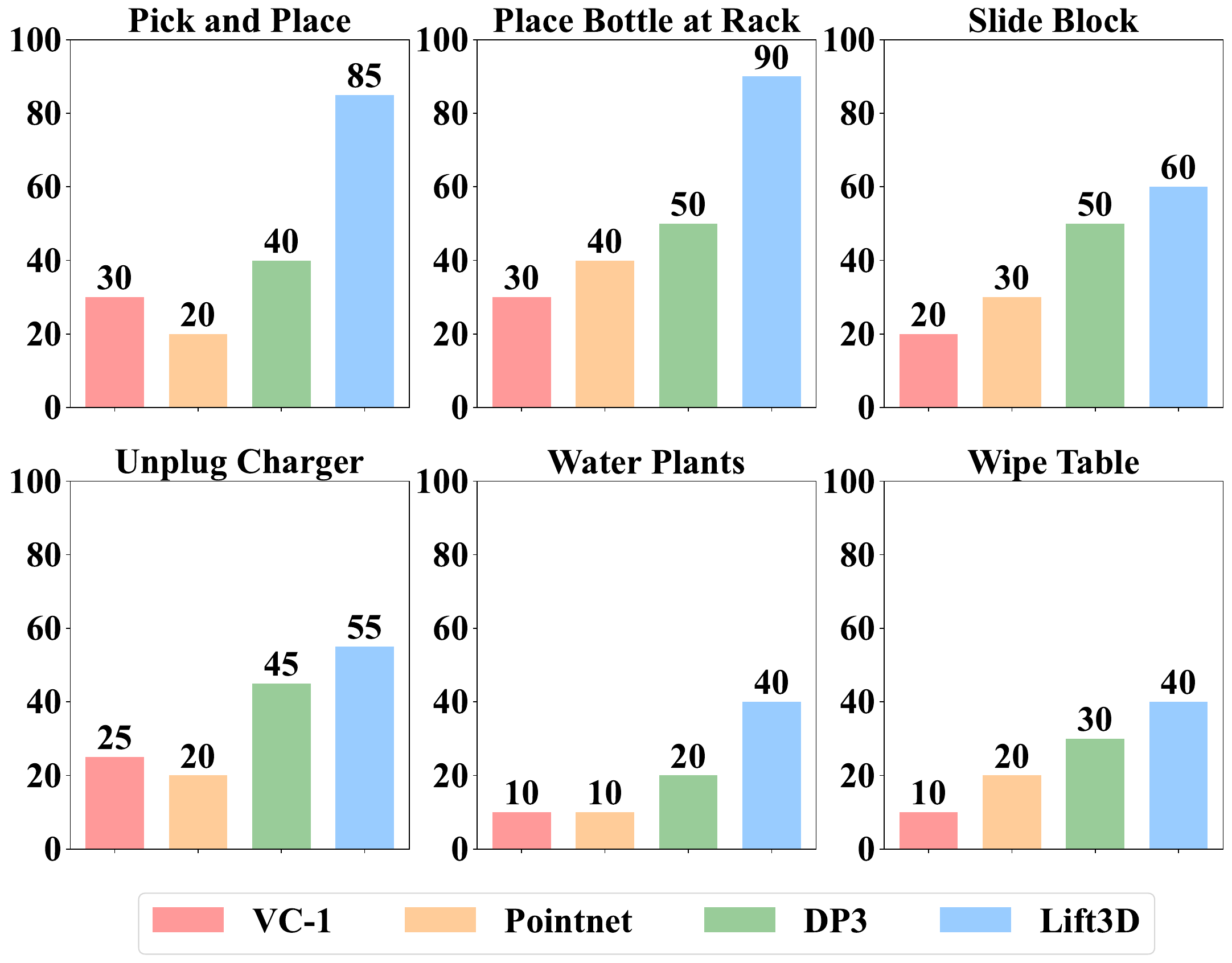}
\vspace{-0.6cm}
\centering
\caption{
\textbf{Quantitative results for real robot experiments}, where the y-axis represents the manipulation success rate.
}
\label{fig:realexp}
\vspace{-0.5cm}
\end{figure}

\noindent \textbf{Quantitative Results.} 
As shown in Figure~\ref{fig:realexp}, we compare Lift3D(CLIP) with DP3~\cite{ze20243d}, VC-1~\cite{majumdar2023we}, and Pointnet~\cite{qi2017pointnet}. The results demonstrate that Lift3D performs consistently well across multiple tasks. Specifically, in the \textit{place bottle at rack} task, which requires accurate 3D position and rotation prediction, Lift3D achieves a 90 success rate. The results demonstrate that Lift3D can effectively understand 3D spatial scenes and make accurate pose predictions in the real world.
Meanwhile, for complex tasks (\textit{wipe table}), all methods face limitations in precision due to the need to manipulate deformable tissue. Despite this, Lift3D still achieves a 40 success rate. Due to space limitations, we place the other real-world experiments in Appendix~\ref{apsec: ARE}.

\noindent \textbf{Qualitative Results.}
As shown in Figure~\ref{fig:realvis}, we visualize the manipulation process of six real-world tasks. Our method accurately predicts the continuous 7-DoF end-effector poses, allowing tasks to be completed along the trajectories. For example, in the \textit{water plants} task, Lift3D first accurately grasps the handle of the watering can. It then smoothly lifts the can and positions it above the plants. Finally, the gripper is gradually rotated to control the flow of ``water".
The demonstration video is provided in the supplementary material, and failure cases are analyzed in Appendix~\ref{apsec: AFC}.

\begin{figure*}[t]
\centering
\includegraphics[width=0.99\linewidth]{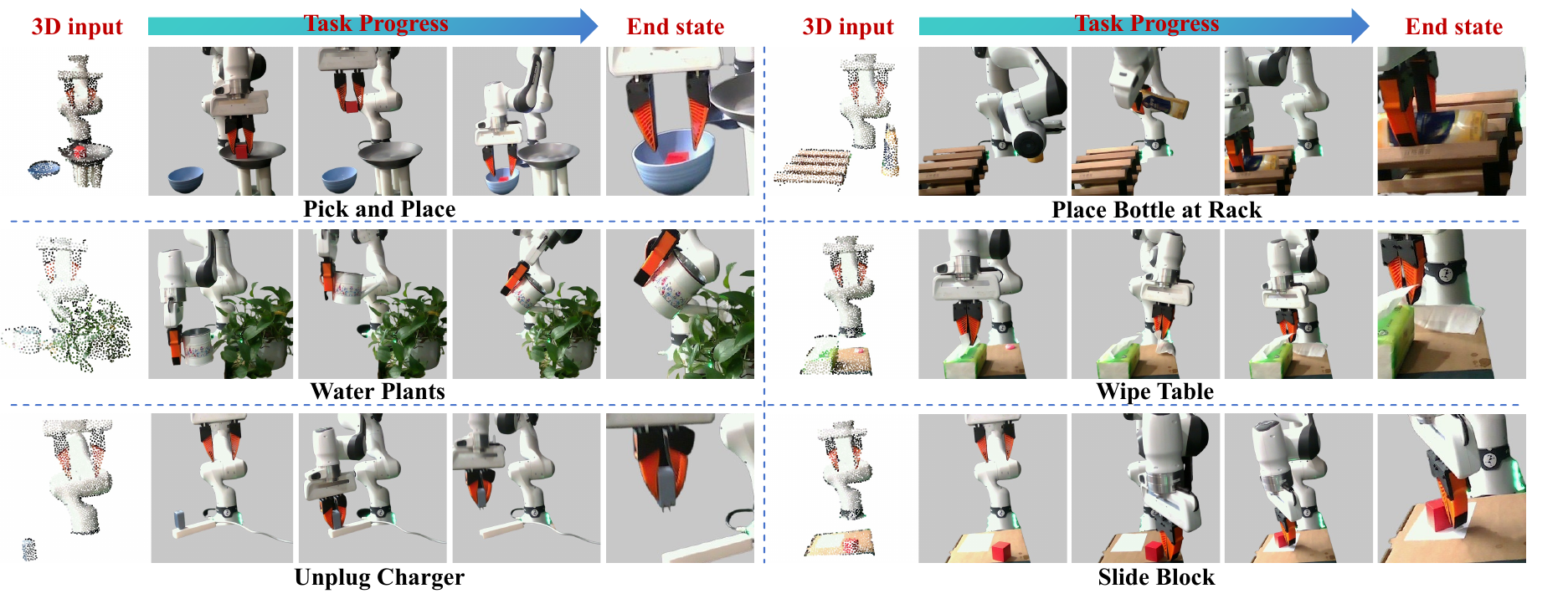}
\vspace{-0.2cm}
\caption{The qualitative results of Lift3D in real-world experiments, including the input point cloud examples, manipulation progress, and the task completion end state, are shown. More visualizations can be found in Appendix~\ref{apsec: AQE2}.
}
\label{fig:realvis} 
\vspace{-0.2cm}
\end{figure*}

\input{tab/table_02_ablation}

\subsection{Ablation Study}
\label{sec: AS}

In Table \ref{tab: abla}, we conduct a series of ablation experiments on 2 MetaWorld simulation tasks, including \textit{assembly} and \textit{box-close}, and calculate the average manipulation accuracy.
\textbf{For the Task-aware MAE}, in Ex2 - Ex4, we observe that Depth and RGB+Depth reconstruction outperform Ex1 with success rates of 6 and 5, respectively, while RGB reconstruction alone does not show a significant improvement. This highlights the importance of reconstructing geometric information in manipulation tasks, leading us to choose Depth as our reconstruction target.
By comparing Ex2 and Ex5, we find that the affordance-guided masking strategy increases the success rate by 4 compared to the random masking strategy, demonstrating that focusing on task-relevant affordance regions for learning geometric information is more efficient.
Compared to Ex5, pretraining with visual token distillation (Ex6) results in an additional increase of 8, suggesting that preventing catastrophic forgetting of pretrained knowledge is essential when endowing 2D foundation models with implicit 3D robotic awareness.
\textbf{For the 2D model-lifting strategy}, compared to Ex1 with image input, Ex7 introduces our lifting strategy with explicit point cloud encoding, which results in a significant improvement. The results demonstrate that 3D spatial information is crucial for achieving robust manipulation.
Ex8 also shows a clear improvement over Ex7, verifying that our implicit 3D representation learning can facilitate subsequent explicit 3D imitation learning.
Finally, compared to Ex8, Ex9 adopts the newly introduced PEs without pretraining and shows a 6 performance decrease, validating that our lifting strategy most effectively leverages large-scale 2D pretrained knowledge.
In Appendix~\ref{apsec: AAS}, we explore the impact of the number and position of virtual planes, and examine the effect of parameter update manner in imitation learning.

\subsection{Exploration of Generalization}
\label{sec: Ge}

\begin{table}[t]
\centering
\setlength\tabcolsep{2.5pt}
\includegraphics[width=0.4\textwidth]{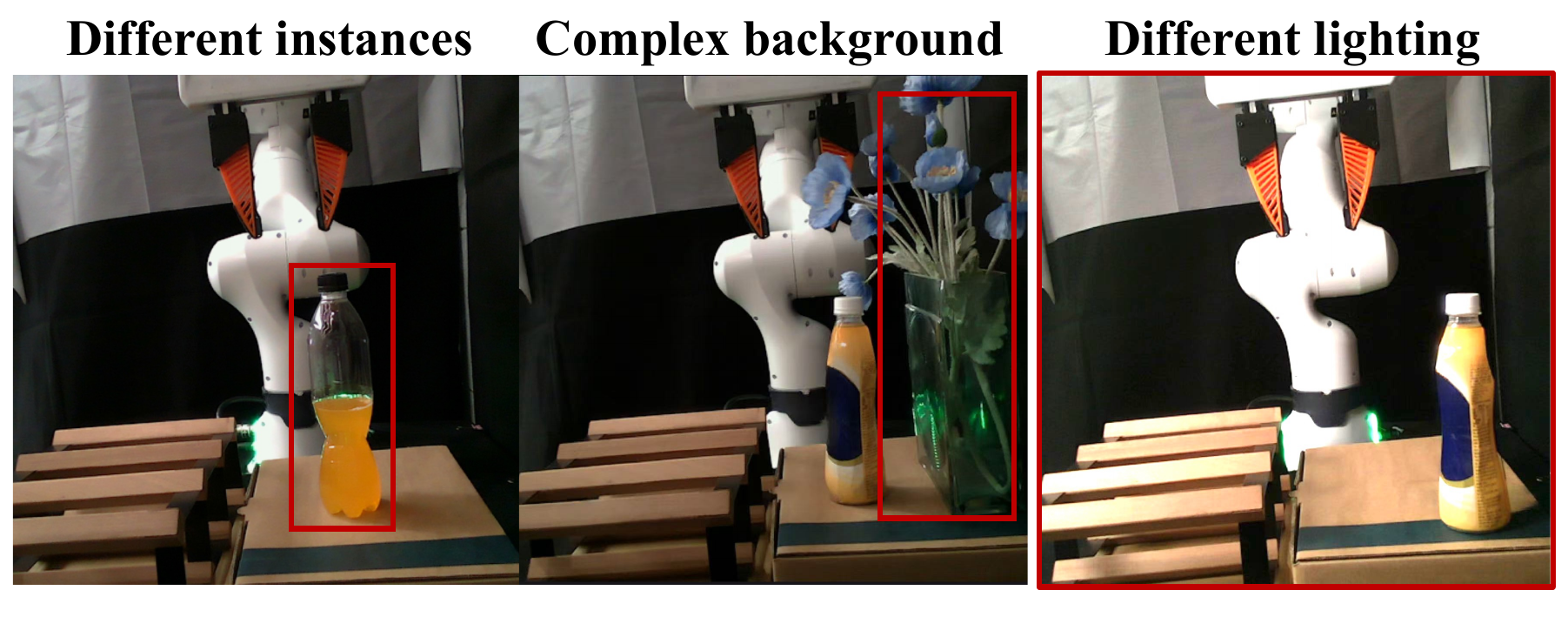}
\scriptsize
\begin{tabular}{c|ccc|ccc}
\hline
Task & \multicolumn{3}{c|}{Pick and place} & \multicolumn{3}{c}{Place bottle at rack}   \\\hline
Scenario & Lift3D & DP3 & VC-1 & Lift3D  & DP3 & VC-1 \\\hline
Original & 85 & 40 & 30 &90 & 50& 30\\
Object & 70(-18\%) & 20(-50\%)& 20(-33\%)& 80(-11\%)& 30(-40\%)& 0(-100\%)\\
Background & 50(-41\%) & 20(-50\%) & 0(-100\%) & 55(-39\%) & 20(-60\%)& 0(-100\%)\\
Brightness & 60(-29\%)& 25 (-37\%)& 20(-33\%) & 80(-11\%) & 40(-20\%) & 20(-33\%)\\
\hline
\end{tabular}

\vspace{-0.3cm}
\caption{\textbf{Generalization.} `Object', `Background', and `Brightness' represent different manipulated objects, background scenes, and lighting conditions, respectively.
The image above illustrates the three test scenarios, with the red box highlighting the main changes.
}
\vspace{-0.4cm}
\label{tab: gen}
\end{table}

Leveraging the large-scale pretrained knowledge of the 2D foundation model and comprehensive 3D robotic representations, Lift3D exhibits strong real-world generalization. 
As shown in Table~\ref{tab: gen}, we design three real-world test scenarios, different from the training scenarios, to validate its generalization ability.
\textbf{1) Different manipulated instances.}
Lift3D demonstrates robustness across various manipulated objects, achieving the smallest accuracy loss. This success can primarily be attributed to the semantic understanding capabilities of the pretrained 2D foundation models.
\textbf{2) Complex background scenes.}
Background distractions significantly reduce accuracy across all methods, but Lift3D shows the smallest drop, maintaining a manipulation success rate above 50.
This can be attributed to the effective use of large-scale pretrained knowledge in 3D space. Additionally, the affordance-guided masking strategy enhances the model's understanding of the spatial geometry of the foreground region through reconstruction, while minimizing the impact of background distractions.
\textbf{3) Different lighting conditions.}
Lighting variations affect the data distribution of 2D images and also impact depth capture, thereby influencing the point cloud data. Under the effect of lighting changes, Lift3D shows only an average accuracy drop of 20\%, demonstrating its robust 3D robotic representation.

\subsection{Exploration of Model Scalability}
\label{sec: MS}

\begin{figure}[t]
\includegraphics[width=0.475\textwidth]{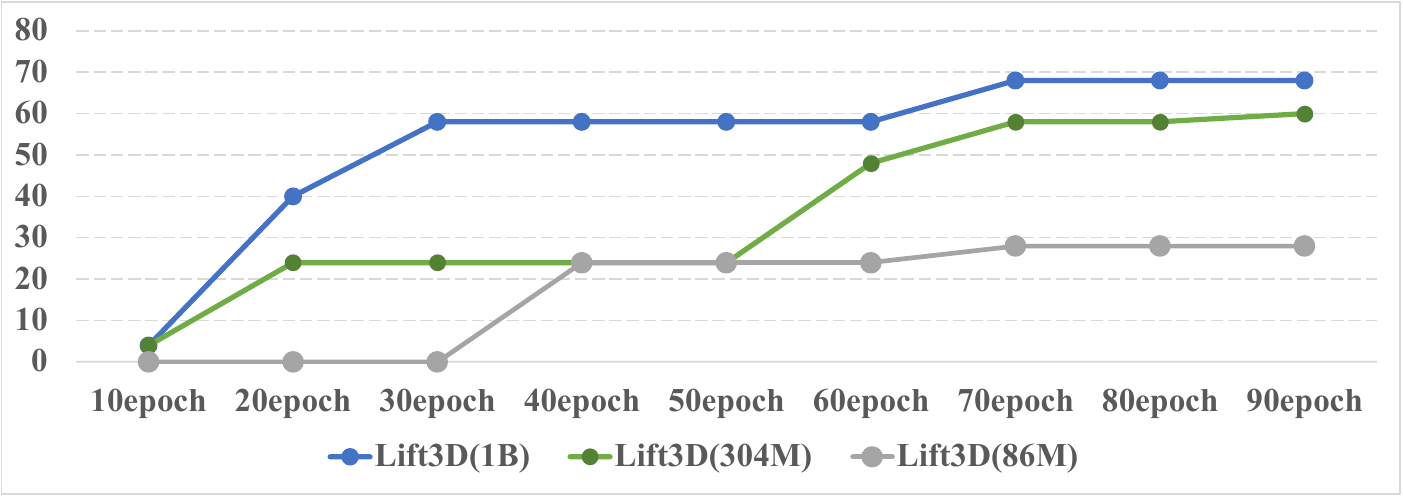}
\vspace{-0.6cm}
\centering
\caption{
\textbf{Scalability.} Y-axis is the manipulation success rate.
}
\label{fig:scale}
\vspace{-0.3cm}
\end{figure}

In computer vision, 2D foundation models typically enhance performance on downstream tasks as their parameters scale up~\cite{oquab2023dinov2, radford2021learning}. Building on this, we investigate whether our proposed Lift3D policy also exhibits scalability. We conduct experiments on the very hard MetaWorld simulation task: \textit{shelf-place}. For this complex task, Lift3D (DINOV2-ViT-base) achieves only 28 accuracy. The parameter count of ViT-base is only 86M, while ViT-large and ViT-giant have 304M and 1B parameters, respectively. By substituting the 2D foundation model with DINOV2-ViT-large and DINOV2-ViT-giant, Lift3D achieves 48 and 58 accuracy on the \textit{shelf-place} task and demonstrates faster convergence, as shown in Figure \ref{fig:scale}. These improvements demonstrate the scalability of the Lift3D policy model, and the Lift3D framework is able to generate more robust manipulation policies with larger 2D foundation models.

\section{Conclusion and Limitation}
In this paper, we introduce Lift3D, a novel framework that integrates large-scale pretrained 2D foundation models with robust 3D manipulation capabilities. First, we design a task-aware MAE that masks task-relevant affordance regions and reconstructs depth geometric information, enhancing implicit 3D robotic representation. Second, we propose a 2D model-lifting strategy that utilizes the pretrained 2D foundation model to explicitly encode 3D point cloud data for manipulation imitation learning.
Lift3D consistently outperforms existing methods in both simulation and real-world experiments, showing strong generalization ability in diverse real-world scenarios.
In terms of limitations, our Lift3D framework focuses on lifting 2D vision models to 3D manipulation tasks, which means it cannot comprehend language conditions. However, our approach can adapt to multimodal models like CLIP, enabling the integration of the Lift3D encoder with a language model and paving the way for a new 3D Vision-Language-Action model.

%% file: tab/table_04_overall.tex
\begin{table*}[t]
\centering
\label{table: simexp}
\resizebox{0.98\textwidth}{!}{%
\begin{tabular}{l|c|c|cccc|c|ccc|c}
\hline
   &   &   & \multicolumn{4}{c|}{MetaWorld} & &\multicolumn{3}{c|}{Adroit} &  \\
Method & Type & Input Type & Easy (7) & Medium (5) & Hard (2) & Very Hard (1) & \textbf{Mean S.R.} & Hammer & Door & Pen & \textbf{Mean S.R.}\\
\hline
CLIP~\cite{radford2021learning} & \multirow{3}{*}{2D Rep.} & RGB & 72.9 & 68.8 & 50.0 & 26.0 & 65.3 & 100 & 100 & 52 & 84.0\\
R3M~\cite{nair2022r3m} &  & RGB & 84.6 & 66.4 & 83.0 & 36.0 & 75.1 & 100 & 100 & 56 & 85.3  \\           
VC-1~\cite{majumdar2023we} &  & RGB & 62.6 & 71.6 & 52.0 & 12.0 & 60.8 & 88 & 100 & 48 & 78.7 \\
\hline
PointNet~\cite{qi2017pointnet} & \multirow{4}{*}{3D Rep.} & PC & 72.0 & 37.6 & 66.0 & 14.0 & 55.9 & 60 & 100 & 48 & 69.3 \\
PointNet++~\cite{qi2017pointnet++} &  & PC & 70.3 & 59.6 & 61.0 & 12.0 & 61.6 & 68 & 100 & 60 & 76.0 \\
PointNeXt~\cite{qian2022pointnext} &  & PC & 82.6 & 62.8 & 59.0 & 20.0 & 68.7 & 52 & 96 & 48 & 65.3 \\
SPA~\cite{zhu2024spa} &  & RGB & 72.6 & 77.4 & 66.0 & 16.0 & 69.5 & 100 & 100 & 44 & 81.3 \\
\hline
DP3~\cite{ze20243d} & \multirow{1}{*}{3D Policy} & PC & 85.7 & 49.6 & 57.0 & 18.0 & 65.3 & 88 & 100 & 12 & 66.7 \\
\hline
\textbf{Lift3D (DINOV2)} & \multirow{2}{*}{Ours} & PC & 93.1 & \textbf{82.4} & \textbf{88.0} & 28.0 & \textbf{84.5} & 100 & 100 & 56 & 85.3 \\
\textbf{Lift3D (CLIP)} &  & PC & \textbf{94.0} & 78.8 & 82.0 & \textbf{42.0} & 83.9 & 100 & 100 & \textbf{64} & \textbf{88.0} \\
\hline
\end{tabular}}
\vspace{-0.2cm}
\caption{\textbf{Comparison of manipulation success rates between Lift3D and 2D \& 3D baselines in simulation benchmarks.} `2D Rep.' and `3D Rep.' refer to robotic 2D representation and 3D representation methods, respectively. `PC' indicates that the model input is point cloud.}
\vspace{-0.3cm}
\label{tab:sim_exp}
\end{table*}

%% file: tab/table_02_ablation.tex








\begin{table}[t]
\centering
\setlength\tabcolsep{4.5pt}
\scriptsize
\begin{tabular}{c|cccc|cc|cc}
\hline
 & \multicolumn{4}{c|}{Task-aware MAE} & \multicolumn{2}{c|}{2D Model-Lifting } & \\
\hline
 &AMS & Depth & RGB & VD & 2ML-PEs &  L-PEs & \textbf{Mean}& \textbf{Gain} \\\hline
Ex1& - & - & - & -& -& -&62 & +0\\
Ex2& - & \checkmark & - & -& -& -& 68& +6\\
Ex3& - & -&\checkmark & -& -& -& 63& +1\\
Ex4& -& \checkmark&\checkmark & -& -& -& 67 & +5 \\
Ex5& \checkmark & \checkmark&- & -& -& -& 72 & +10\\
Ex6& \checkmark & \checkmark&- & \checkmark& -& -& 80 & +18\\
Ex7& - & -&- & -& \checkmark& -& 86 & +14\\
Ex8& \checkmark & \checkmark&- & \checkmark& \checkmark& -&96 & +34\\
Ex9& \checkmark & \checkmark&- & \checkmark& -& \checkmark&90 & +28\\
\hline
\end{tabular}
\vspace{-0.3cm}
\caption{\textbf{Ablation study.} In the Task-aware MAE, AMS refers to the affordance-guided masking strategy, Depth and RGB refer to the reconstruction targets, and VD stands for visual token distillation. In the 2D Model-Lifting, 2ML-PEs and L-PEs represent the pretrained 2D PEs constructed using our method and the newly injected learnable PEs, respectively.}
\vspace{-0.55cm}
\label{tab: abla}
\end{table}

%% file: sec/Appendix.tex
\appendix

Due to space limitations, we provide additional details, as well as quantitative and qualitative results of our Lift3D in this supplementary material. The outline is shown below.

\begin{itemize}
    \item \textbf{A. Additional Details (Appendix~\ref{apsec: AD})}
    \begin{itemize}
        \item Details of Reconstruction Dataset
        \item Additional Details of the Real-World Dataset
    \end{itemize}
    \item \textbf{B. Additional Quantitative Experiments (Appendix~\ref{apsec: AQE1})}
    \begin{itemize}
        \item RLBench Experiments
        \item Additional Real-World Experiments
        \item Detail score of MetaWorld
        \item Additional Ablation Study
    \end{itemize}
    \item \textbf{C. Additional Qualitative Experiments (Appendix~\ref{apsec: AQE2})}
    \begin{itemize}
        \item Additional Real-World Visualization
        \item Additional Failure Case Analysis
    \end{itemize}
\end{itemize}

\section{Additional Details}
\label{apsec: AD}
Our training dataset is divided into two parts, systematically empowering the 2D foundation model with 3D robotic manipulation capabilities. In Sections~\ref{apsec: DRD}, we provide additional details of the reconstruction dataset, which is used in implicit 3D robotic representations pretraining. In Sections~\ref{apsec: DRWD}, we provide additional details of the real-world dataset, which is used in explicit 3D imitation learning.

\subsection{Details of Reconstruction Dataset}
\label{apsec: DRD}

Since most subsets in the open x-embodiment dataset~\cite{o2023open} do not simultaneously contain both camera parameters and depth, we are unable to construct point cloud data for our explicit 3D imitation learning (stage 2). Therefore, we leverage this dataset to build our MAE training data. First, we select subsets that contain paired RGB, depth, and text description data. Second, we randomly sample one frame from every nine frames in each episode. As a result, the reconstruction dataset provides 1 million image-depth-text pairs. The images are used as model input, depth serves as the reconstruction target, and the text descriptions are used for task-related affordance generation. The selected subsets are:

\begin{itemize}
    \item \textit{tacoplay}
    \item \textit{berkeley\_autolab\_ur5}
    \item \textit{uiuc\_d3field}
    \item \textit{nyu\_franka\_play\_dataset\_converted\_externally\_to\_rlds}
    \item \textit{stanford\_robocook\_converted\_externally\_to\_rlds}
    \item \textit{maniskill\_dataset\_converted\_externally\_to\_rlds}
\end{itemize}

\subsection{Additional Details of the Real-World Dataset}
\label{apsec: DRWD}

\begin{figure}[ht!]
\centering
\includegraphics[width=\linewidth]{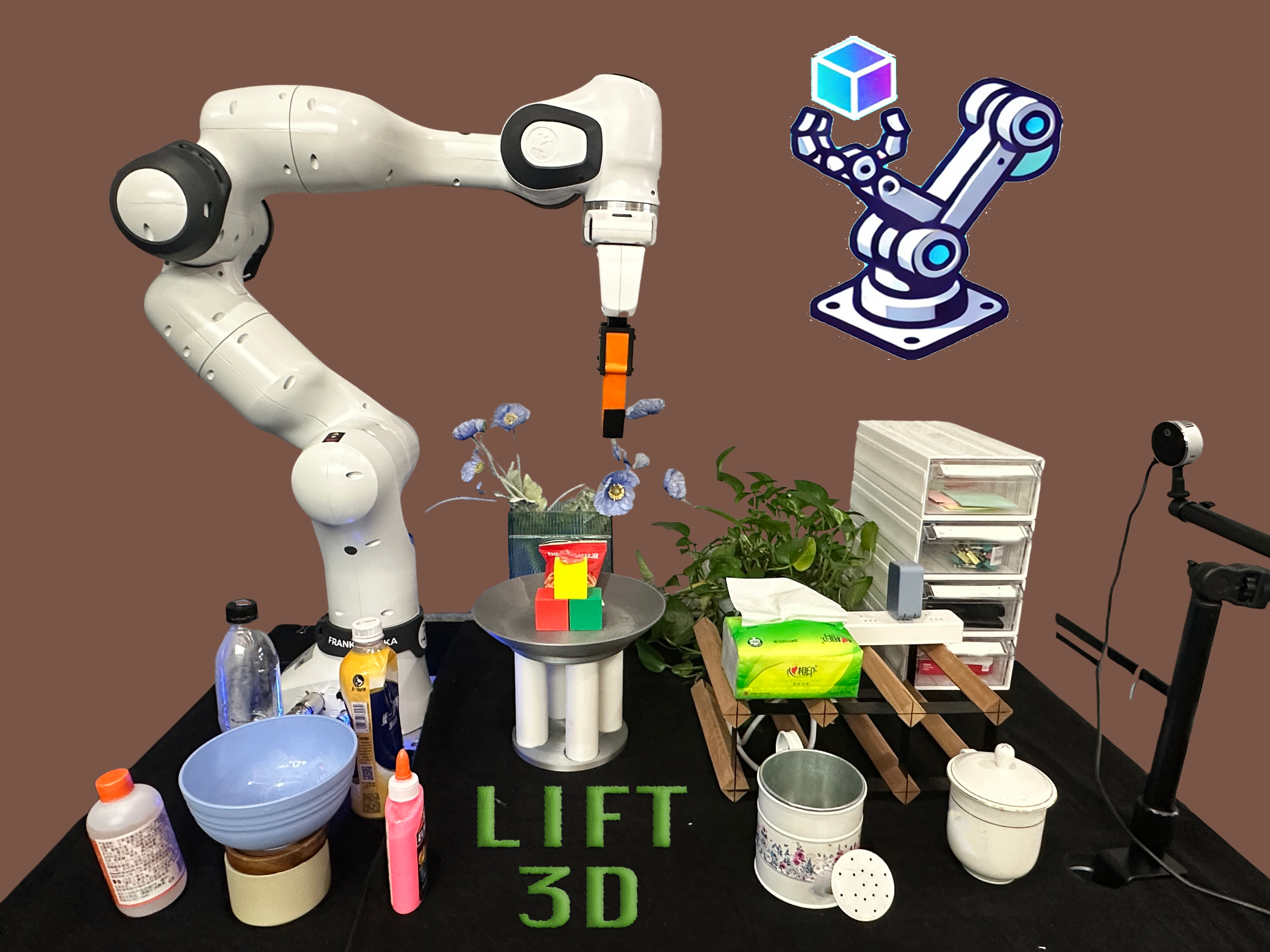}
\caption{\textbf{Real-world scenario.} The assets and environment configured for the real-world experiments.}
\label{apfig:asset} 
\end{figure}

We use the Franka Research 3 (FR3) arm as the hardware platform for our real-world experiments. Due to the relatively short length of the FR3 gripper fingers, which makes it challenging to perform certain complex tasks, we 3D print and replace the original gripper with a UMI gripper~\cite{chi2024universal}. We conduct ten tasks, selecting 30 episodes and extracting key frames to construct the training set for each task. The number of key frames per task varies, as follows: \textbf{3} frames for \textit{unplug charger}, \textit{slide block}, \textit{open drawer}, \textit{close drawer}; and \textbf{4} frames for \textit{place bottle at rack}, \textit{pour water}, \textit{pick and place}, \textit{water plants}, \textit{wipe table}. 
The experimental assets and environment are shown in Figure~\ref{apfig:asset}.
During the evaluation of real-world tasks, we determine the success of each task based on human assessment. The successful states of the 10 tasks are shown in the End State images in Figure 4 of the main text and Figure~\ref{apfig:realvis} of the appendix.

\section{Additional Quantitative Experiments}
\label{apsec: AQE1}

In Section~\ref{apsec: RE}, we compare our method against other baselines using the RLBench simulator benchmark. Additional real-world experiments are presented in Section~\ref{apsec: ARE}, which include four real-world tasks not covered in the main text. The fine-grid success rates for each task in the MetaWorld benchmark are provided in Section~\ref{apsec: DSM}. Finally, Section~\ref{apsec: AAS} investigates the impact of the number and positioning of virtual planes, evaluates the effect of parameter update strategies, and analyzes the influence of the 3D tokenizer’s parameter size on 3D imitation learning.

\subsection{RLBench Experiments}
\label{apsec: RE}

\begin{table*}[t]
\centering
\setlength\tabcolsep{2pt}
\small
\begin{tabular}{c|c|cccccc|c}
\hline
Method& Input Type & Close box & Put rubbish in bin & Close laptop lid &Water plants &Unplug charger &Toilet seat down& Mean \\
\midrule
VC-1~\cite{majumdar2023we}& Single-view RGB & 52   & 12  & 88  &  12 &  28 &  96 & 48.0 \\
PointNet~\cite{qi2017pointnet}& Single-view PC & 52  &  56  &  88 &  20 & 36 & 96 &  58.0 \\
RVT-2~\cite{goyal2024rvt}& Four-view PC & 88  & 100  & \textbf{100} &12& 4& \textbf{100} & 67.3  \\

RVT-2~\cite{goyal2024rvt}& Single-view PC &  \textbf{96} & \textbf{100}  & 76 & 16 & 8 & 96& 65.3  \\
\textbf{Lift3D} & Single-view PC & 92 & 80 & 92 & \textbf{36}& \textbf{36} & \textbf{100} & \textbf{72.6}  \\
\hline
\end{tabular}
\caption{\textbf{Comparison of manipulation success rates between Lift3D and 2D \& 3D baselines in RLBench benchmarks.}
`Single-view PC' and `Four-view PC' indicate the use of one or four different viewpoints of RGBD cameras to construct the input point cloud data, which does not indicate the number of virtual planes in RVT-2.
}
\label{aptab: rlbench}
\end{table*}

\input{tab/table_07_realworld_appendix}

\textbf{Experiment setting.} In the RLBench benchmark~\cite{james2020rlbench}, the data are collected through pre-defined waypoints and the Open Motion Planning Library~\cite{sucan2012open}. Each task consists of 100 gathered episodes. Following previous work~\cite{shridhar2022peract, goyal2023rvt, goyal2024rvt}, we use key frames to construct the training dataset. For baseline comparison, we select VC-1~\cite{majumdar2023we}, PointNet~\cite{qi2017pointnet}, and RVT-2~\cite{goyal2024rvt}. Since Lift3D and PointNet require only single-view point cloud data as input, we compare RVT-2 in two settings: using single or four different viewpoints of RGBD cameras to construct the input point cloud data. Note that, many existing policies~\cite{ze20243d, gervet2023act3d} use single-view point cloud as input, which is a more practical and low-cost approach for real-world applications. The training details are consistent with the simulation experiments described in Section 4.1 of the main text. For a fair comparison, we ensure all methods have the same model throughput and train for the same number of iterations.

\textbf{Quantitative Results.} 
In Table~\ref{aptab: rlbench}, Lift3D(CLIP) achieves an average success rate of 72.6 on RLBench. Compared to 2D and 3D robotic representation methods, Lift3D improves the mean success rate by 24.6 and 14.6, respectively. These results demonstrate that our method effectively enhances the 3D robotic representation of the 2D foundation model.
Meanwhile, when comparing Lift3D to RVT-2 under single-view point cloud input, Lift3D achieves an accuracy improvement of 7.3. Even when compared with RVT-2 using four-view point cloud input, Lift3D still achieves comparable results. 
These findings indicate that even with single-view point cloud data, our method demonstrates robust manipulation capabilities, highlighting the strong practicality of Lift3D.
Unlike RVT-2, our Lift3D model does not utilize a language model or language prompts for task differentiation. Instead, it exclusively relies on point clouds and robot states as input.
In future work, we plan to enhance our framework by incorporating a language model to better handle human language conditions, which is highly feasible and straightforward. For example, we integrate a CLIP-BERT~\cite{radford2021learning} model into Lift3D(CLIP) for language text encoding.

\subsection{Additional Real-World Experiments}
\label{apsec: ARE}

\input{tab/table_06_app_metaworldall}

\input{tab/table_05_app_ablation}

As shown in Table~\ref{aptab: real}, we present the results for the remaining four real-world tasks.
\textbf{Pour Water}: This challenging task requires complex action predictions and precise gripper rotation for controlling the bottle. Lift3D achieves a success rate of 85\%, representing a 25\% improvement over the previous SOTA (DP3). Lift3D successfully completes the sequence of bottle grasping, bottle moving, and precise rotation, demonstrating its significant potential in handling complex tasks.
\noindent\textbf{Stack Blocks}: This task demands precise spatial understanding and position prediction. Lift3D accurately predicts the positions of both the grasping and placement blocks, achieving a 35\% success rate. While the accuracy is not optimal, it still outperforms other methods, demonstrating superior 3D robotic representation.
\noindent\textbf{Open/Close Drawer}: These tasks assess the model's ability to interact with articulated objects. Lift3D achieves success rates of 60\% and 75\% for opening and closing drawers, respectively. It accurately predicts the grasp position and rotation for the drawer handle, as well as the precise trajectory for the opening and closing motion. The results demonstrate that Lift3D can not only predict accurate 6-DoF poses but also predict motion trajectories for articulated objects.
Based on all real-world results, we evaluate the exceptional 3D robotic representation and pretraining knowledge of our Lift3D policy, which demonstrates robust manipulation capabilities across diverse real-world tasks, even with only 30 episodes of training data.

\subsection{Detail Score of MetaWorld}
\label{apsec: DSM}
As shown in Table~\ref{table-ap:simexp2}, we present the fine-grid scores for each task in MetaWorld. The reported scores represent the average success rate across two camera views: Corner and Corner2. Lift3D (CLIP) ranks first in 8 tasks with an average success rate of 83.9\%, while Lift3D (DINOv2) ranks first in 11 tasks with an average success rate of 84.5\%. Notably, Lift3D (DINOv2) achieves nearly 100\% success in 7 tasks, and Lift3D (CLIP) does so in 5 tasks. These results demonstrate that Lift3D effectively enhances both the implicit and explicit 3D robotic representations of 2D foundation models, regardless of their pretraining methods.
However, on the \textit{push-wall} task, Lift3D does not achieve leading performance. By visualizing the model's input, we find that the sparsity of the point cloud on the wall leads to inaccuracies in predicting the push position. In future work, we plan to increase the density of the input point cloud, enabling the model to extract more precise and detailed explicit 3D robotic representations.

\subsection{Additional Ablation Study}
\label{apsec: AAS}

\begin{figure*}[ht!]
\centering
\includegraphics[width=\linewidth]{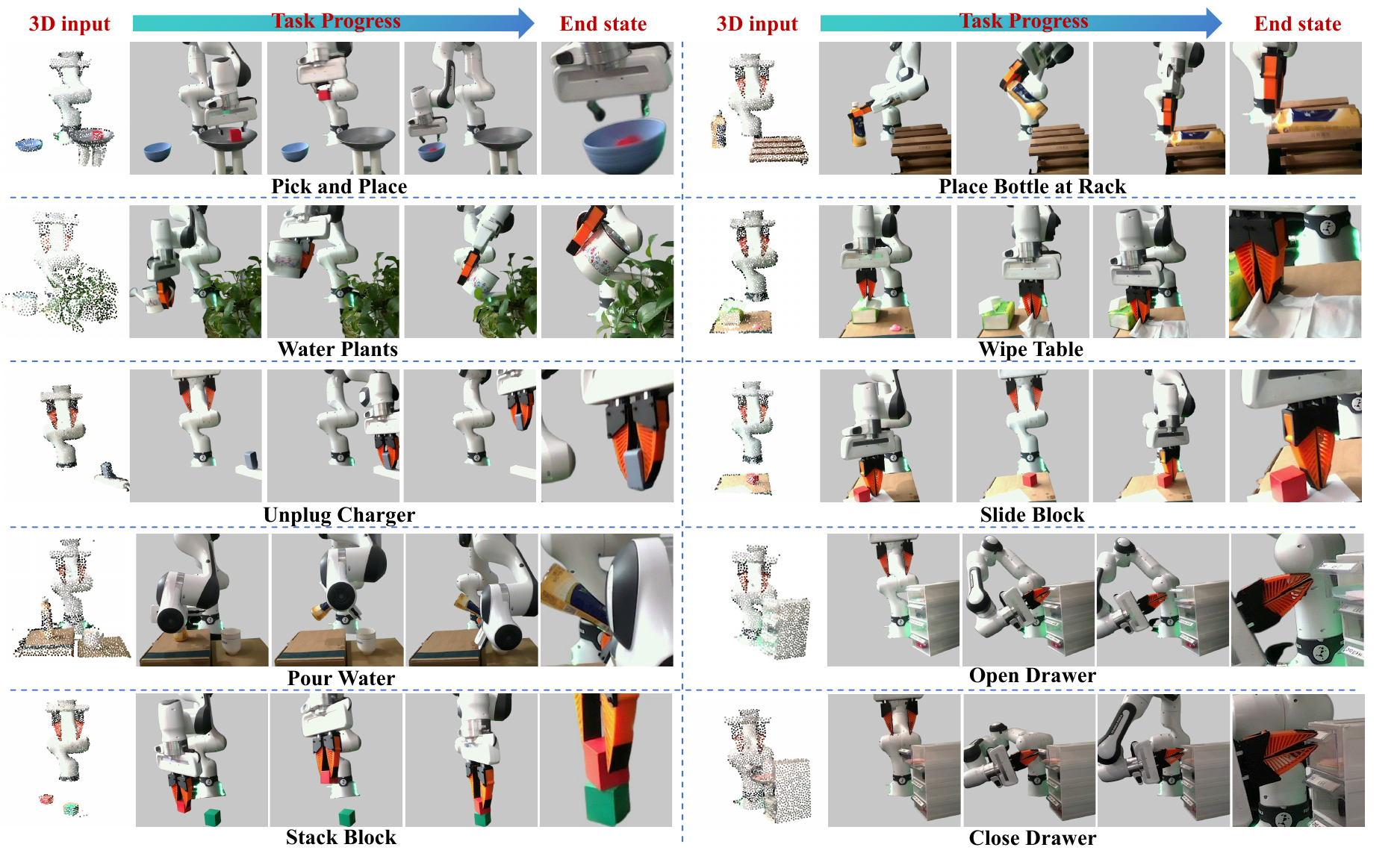}
\caption{\textbf{The qualitative results of Lift3D in real-world experiments}, including the input point cloud examples, manipulation progress, and the task completion end state, are shown. The visualization case differs from the samples presented in the main text.
}
\label{apfig:realvis} 
\end{figure*}

In Table~\ref{ap-tab: abla}, we present three additional ablation study on the Metaworld benchmark, which use the same task of main text (\textit{assembly} and \textit{box-close}), reporting the average success rate.

\noindent \textbf{Number of Virtual Planes.} We analyze the effect of varying the number and positions of virtual planes, which are used for positional mapping between the 3D input points and the 2D positional embeddings. The main paper reports results using six planes (top, bottom, left, right, front, and back). Here, we compare this setup with configurations using four planes (front, back, left, and right), two planes (front and back), and a single plane (front). The results, as shown in Table~\ref{ap-tab: abla}, indicate that the six-plane configuration achieves the best performance (96\%), followed by two planes (92\%), four planes (88\%), and one plane (86\%). This demonstrates that using six planes provides diverse positional relationships from multiple perspectives, better encodes the positional information of point cloud data, and minimizes the loss of spatial information.

\noindent \textbf{Parameter Update Strategy for Imitation Learning.} We evaluate the impact of different parameter update strategies during the imitation learning stage (Stage 2). For all strategies, we update the 3D tokenizer and policy head, both of which are randomly initialized. Our proposed method injects LoRA~\cite{hu2021lora} into foundation models for parameter-efficient fine-tuning~\cite{liu2023vida, yang2023exploring}. Specifically, we explore two configurations: (a) full fine-tuning of all parameters and (b) excluding LoRA injections. The results, summarized in Table~\ref{ap-tab: abla}, indicate that our adopted update strategy achieves the highest mean success rate (96\%), though the performance differences are minor. These findings suggest that when the policy has strong 3D robotic representations, it can deliver robust manipulation regardless of parameter update strategies. Meanwhile, Lift3D's update strategy is highly efficient, updating only 1.01M parameters—just 1\% of the total model.

\noindent \textbf{Number of Layers in the 3D Tokenizer.} 
Different layers used in the 3D tokenizer affect the module's parameter size. Our method employs a 3-layer 3D tokenizer specifically designed for efficiently converting point clouds into 3D tokens. Each layer integrates Farthest Point Sampling~\cite{qi2017pointnet} to reduce the number of points, the k-Nearest Neighbor algorithm (k=64) for local feature aggregation, and learnable linear layers for feature encoding. The main paper presents results based on the 3-layer configuration.
Additionally, we conduct experiments comparing setups with 1, 2, 3, and 4 layers, where the token feature channel dimensions are set to 192, 384, 768, and 1536, respectively. For varying dimensions, we incorporate an additional linear layer to align the channel dimensions with those required by the 2D foundation model (e.g., CLIP-ViT-base uses a channel dimension of 768).
As shown in Table~\ref{ap-tab: abla}, the 3-layer configuration achieves the highest success rate (96\%), outperforming the others: 4 layers (94\%), 2 layers (90\%), and 1 layer (76\%). While both the 3-layer and 4-layer configurations demonstrate strong performance, the 3-layer setup emerges as the optimal choice, offering a better balance between accuracy and computational efficiency.

\section{Additional Qualitative Experiments}
\label{apsec: AQE2}

\begin{figure*}[t]
\centering
\includegraphics[width=0.85\linewidth]{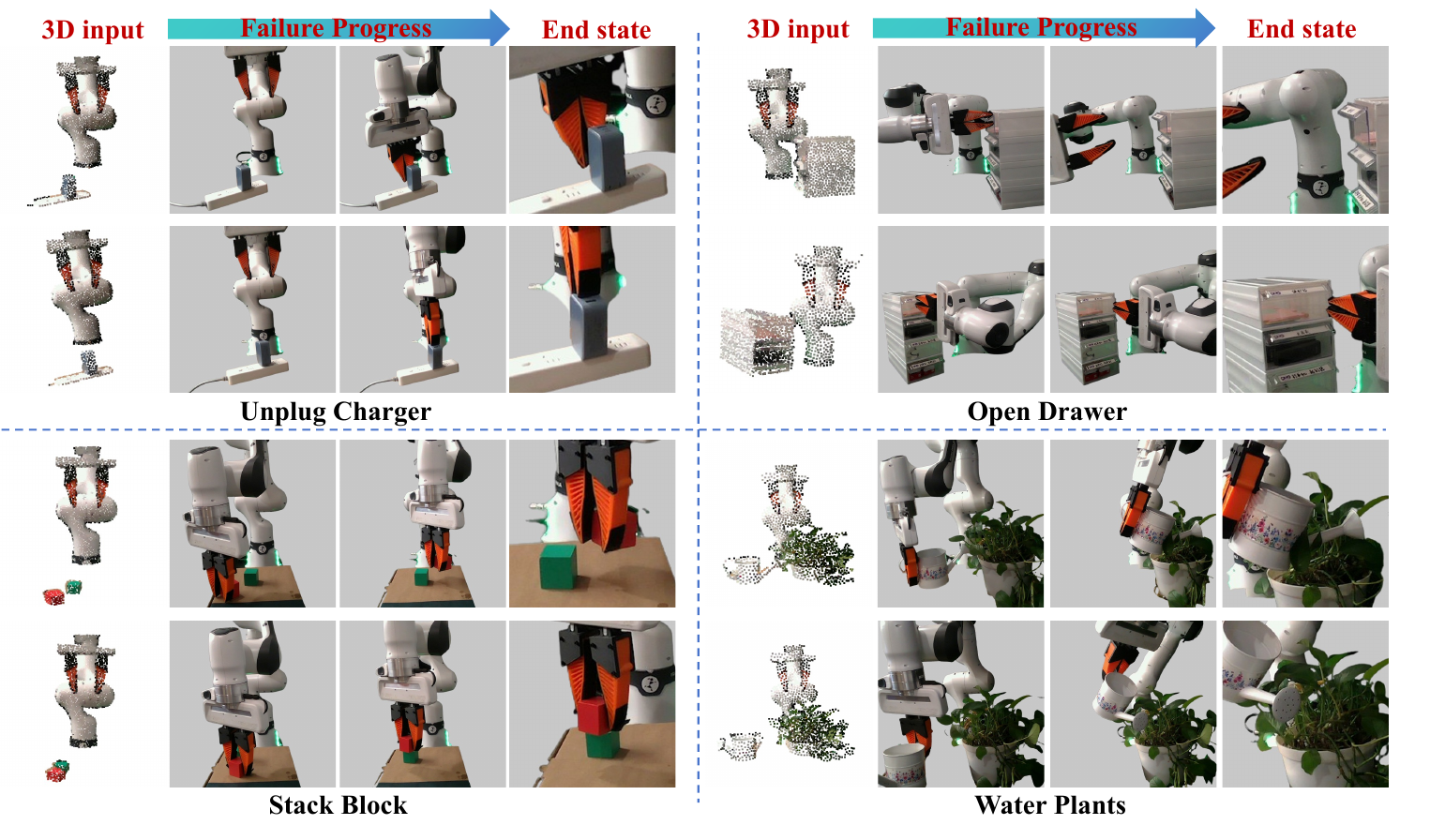}
\caption{\textbf{The failure cases of Lift3D in real-world experiments}, including examples of input point clouds, manipulation progress, and the failure end states, are presented. The tasks include \textit{unplug charger}, \textit{open drawer}, \textit{stack block}, and \textit{water plants}.
}
\label{apfig:failvis} 
\end{figure*}

In Section~\ref{apsec: ARV}, we visualize the manipulation process of four real-world tasks not covered in the main text.
In Section~\ref{apsec: ARV}, we visualize the failure cases in real-world experiments and analyze the failure reasons.

\subsection{Additional Real-World Visualization}
\label{apsec: ARV}

As shown in Figure~\ref{apfig:realvis}, we visualize the manipulation processes of ten real-world tasks.
All visualization results are derived from our proposed Lift3D(CLIP-ViT-base) policy model. Each real-world task is specifically designed to evaluate different capabilities of the Lift3D policy model.
For the first six tasks, we selected visualization cases different from those presented in the main text.
Our method accurately predicts 7-DoF end-effector poses, enabling tasks to be completed along the trajectories. For instance, in the \textit{stack block} task, Lift3D first accurately grasps the red block, lifts it smoothly, and then precisely places it directly above the green block. This task highlights the model's spatial reasoning capabilities, requiring precise perception of the red and green blocks' positions and an accurate understanding of their spatial relationships. 
Demonstration videos are provided in the supplementary material.

\subsection{Additional Failure Case Analysis}
\label{apsec: AFC}
As shown in Figure~\ref{apfig:failvis}, through extensive real-world experiments, we identified four primary categories of failure cases that affect the performance of Lift3D. 
The first category, \textbf{loss of control}, typically occurs during interactions with target objects, such as \textit{open drawer} and \textit{close drawer}. It is characterized by inconsistent force application when handling objects of varying weights and sudden gripper slippage on smooth surfaces.
\textbf{Rotational prediction deviations} constitute the second category of failures, particularly evident in tasks requiring precise rotation control, such as \textit{water plants}, \textit{pour water}, and \textit{place bottle at rack}. These failures include accumulated errors in multi-step rotational movements and incorrect rotation angles when interacting with target objects.
The third category encompasses pose predictions that exceed the robot's degree of \textbf{freedom limits}. The model occasionally predicts poses that exceed the mechanical limits of the Franka robotic arm, generates target positions that are unreachable due to workspace boundaries, or produces kinematically infeasible configurations during complex transitions.

%% file: tab/table_07_realworld_appendix.tex
\begin{table*}[t]
\centering
\setlength\tabcolsep{7pt}
\small
\begin{tabular}{c|c|c|cccc|c}
\hline
Method& Input Type &Reference& Pour Water & Stack Block & Open Drawer &Close Drawer & Mean \\
\midrule
VC-1~\cite{majumdar2023we}& Neurips 2023& RGB &  35 &  0 &  10 &  35 & 20 \\
PointNet~\cite{qi2017pointnet}& CVPR 2017 &PC &  30 &  15  & 30 &  30 & 26 \\
DP3~\cite{ze20243d}& RSS 2024 &PC &  60 & 30  & \textbf{60} &  70 & 55 \\
\textbf{Lift3D} & Ours & PC &\textbf{85} & \textbf{35} & \textbf{60} & \textbf{75} & \textbf{63}  \\
\hline
\end{tabular}
\caption{\textbf{Quantitative results for real robot experiments.} The training setup is consistent with the real-world experiments described in the main text. 
For evaluation, we use the model from the final epoch and test it 20 times across diverse spatial positions.
`RGB' and `PC' indicate that the model input is 2D images and 3D point cloud, respectively.}
\label{aptab: real}
\end{table*}

%% file: tab/table_06_app_metaworldall.tex
\begin{table*}[t]
\centering
\resizebox{0.98\textwidth}{!}{%
\begin{tabular}{l|ccc|c|ccccccccc}
\hline
    & \multicolumn{3}{c|}{Adroit} & {} & \multicolumn{5}{c}{MetaWorld} &  {}\\
Algorithm & Hammer & Door & Pen & \textbf{Mean S.R.} & Button-press & Drawer-open & Reach & Hammer & Handle-pull & Peg-unplug-side \\
\hline
CLIP & 100 & 100 & 52 & 84.0 & 100 & 100 & 56 & 88 & 22 & 78\\
R3M & 100 & 100 & 56 & 85.3 & 92 & 100 & 60 & 60 & 68 & 96 \\           
VC-1 & 88 & 100 & 48 & 78.7 & 96 & 100 & 30 & 88 & 10 & 50 \\
\hline
PointNet & 60 & 100 & 48 & 69.3 & 100 & 100 & 52 & 38 & 4 & 68 \\
PointNet++ & 68 & 100 & 60 & 76.0 & 98 & 84 & 48 & 70 & 12 & 78 \\
PointNeXt & 52 & 96 & 48 & 65.3 & 100 & 100 & 36 & 50 & 78 & 92 \\
SPA & 100 & 100 & 44 & 81.3 & 100 & 96 & 56 & 100 & 36 & 68 \\
\hline
DP3 & 88 & 100 & 12 & 66.7 & 100 & 100 & 40 & 100 & 90 & \textbf{98} \\
\hline
\textbf{Lift3D(Dinov2)} & 100 & 100 & 56 & 85.3 & 100 & 100 & \textbf{80} & 100 & 100 & 96 \\
\textbf{Lift3D(Clip)} & 100 & 100 & \textbf{64} & \textbf{88.0} & 100 & 100 & 74 & 94 & 100 & \textbf{98} \\
\hline
\end{tabular}}

\vspace{0.1in}
\resizebox{0.98\textwidth}{!}{%
\begin{tabular}{l|ccccccccc|c}
\hline
    & \multicolumn{9}{c|}{MetaWorld} \\
Algorithm & Lever-pull & Dial-turn & Sweep-into & Bin-picking & Push-wall & Box-close & Assembly & Hand-insert & Shelf-place & \textbf{Mean S.R.}\\
\hline
CLIP & 72 & 82 & 40 & 92 & \textbf{64} & 60 & 64 & 36 & 26 & 65.3 \\
R3M & 76 & 100 & 60 & 60 & 60 & 92 & 100 & 66 & 36 & 75.1 \\
VC-1 & 76 & 76 & 60 & 80 & 64 & 66 & 60 & 44 & 12 & 60.8 \\
\hline
PointNet & 86 & 94 & 24 & 44 & 36 & 46 & 100 & 32 & 14 & 55.9 \\
PointNet++ & \textbf{94} &  78 & 42 & 72 & 28 & 86 & 96 & 26 & 12 & 61.6 \\
PointNeXt & 80 & 92 & \textbf{78} & 82 & 26 & 78 & 98 & 20 & 20 & 68.7 \\
SPA & 68 & 84 & 64 & 92 & 55 & 76 & 96 & 36 & 16 & 69.5 \\
\hline
DP3 & 80 & 92 & 22 & 24 & 54 & 48 & 100 & 14 & 18 & 65.3 \\
\textbf{Lift3D(Dinov2)} & 76 & 100 & \textbf{80} & \textbf{100} & 40 & \textbf{92} & 100 & \textbf{76}  & 28 & \textbf{84.5} \\
\textbf{Lift3D(Clip)} & 86 & 100 & 72 & 92 & 44 & \textbf{92} & 100 & 64 & \textbf{42} & 83.9 \\
\hline
\end{tabular}}
\caption{\textbf{Comparison of manipulation success rates between Lift3D and 2D \& 3D baselines.} The table presents task-specific scores for each method, covering 18 tasks in Metaworld and 3 tasks in Adroit.}
\label{table-ap:simexp2}
\end{table*}

%% file: tab/table_05_app_ablation.tex
\begin{table}[t]
\centering
\setlength\tabcolsep{5pt}
\small
\begin{tabular}{c|c|c|c}
\hline
\textbf{Experiment}           & \textbf{Configuration} & \textbf{Parameters} & \textbf{Mean} \\ \hline
\multirow{4}{*}{Virtual Planes} 
                              & 1 plane  & -                  & 86                          \\ \cline{2-3} 
                              & 2 planes  & -                  & 92                         \\ \cline{2-3} 
                              & 4 planes   & -                  & 88                         \\ \cline{2-3} 
                              & \textbf{6 planes} & -                     & \textbf{96}                         \\ \hline
\multirow{3}{*}{Update Strategy} 
                              & \textbf{LoRA}  &  1.01M                 & \textbf{96}                         \\ \cline{2-3} 
                              & Without LoRA  &  0.87M          & 90                         \\ \cline{2-3} 
                              & Full Fine-Tuning  &  116.79M      & 92                         \\ \hline
\multirow{4}{*}{3D tokenizer} 
                              & 1 layer     &  0.37M                 & 76                         \\ \cline{2-3} 
                              & 2 layers     &   0.66M                 & 90                         \\ \cline{2-3} 
                              & \textbf{3 layers}  &  1.01M                     & \textbf{96}                         \\ \cline{2-3} 
                              & 4 layers     &  3.96M                 & 94                         \\ \hline
\end{tabular}
\caption{Ablation study results evaluating the influence of (1) the number of virtual planes for positional embeddings, (2) different parameter update strategies (LoRA, without LoRA, and full fine-tuning), and (3) the number of layers in the 3D tokenizer. Success rates highlight the advantages of the proposed configurations}
\label{ap-tab: abla}
\end{table}